\documentclass[12pt,a4paper]{article}
\usepackage[left=1in,right=1in,top=1in,bottom=1in]{geometry}

\usepackage[numbers]{natbib}
\usepackage{float} 
\usepackage[utf8]{inputenc} % allow utf-8 input
\usepackage[T1]{fontenc}    % use 8-bit T1 fonts
\usepackage{hyperref}       % hyperlinks
\usepackage{url}            % simple URL typesetting
\usepackage{booktabs}       % professional-quality tables
\usepackage{amsfonts}       % blackboard math symbols
\usepackage{nicefrac}       % compact symbols for 1/2, etc.
\usepackage{microtype}      % microtypography
\usepackage{lipsum}		% Can be removed after putting your text content
\usepackage{graphicx}
\usepackage{natbib}
\usepackage{doi}
\usepackage{amsmath}  % For math environments and \text command
\usepackage{colortbl}    % For colored table cells (\rowcolor)
\usepackage{xcolor}      % For color definitions
\title{Beyond One-Size-Fits-All Summarization: Customizing Summaries for Diverse Users}

\author{Mehmet Samet Duran\thanks{Mehmet Samet Duran \newline Faculty of Engineering and Natural Sciences, Bahcesehir University\newline Istanbul, Turkey. \newline E-mail: mehmet.duran@bahcesehir.edu.tr}
\and Tevfik Aytekin\thanks{Tevfik Aytekin\newline Department of Computer Engineering, Bahcesehir University, Ciragan Caddesi, 34353 \newline Besiktas, Istanbul, Turkey. \newline Tel.: +90-212-3810580, \newline Fax: +90 212 3810550, \newline E-mail: tevfik.aytekin@bau.edu.tr}}

\date{}

\hypersetup{
pdftitle={READABILITY-AWARE TEXT SUMMARIZATION},
pdfauthor={Mehmet Samet Duran, Assoc. Prof. Tevfik Aytekin},
}

\begin{document}
\maketitle

\begin{abstract}
In recent years, automatic text summarization has witnessed significant
advancement, particularly with the development of transformer-based models.
However, the challenge of controlling the readability level of generated summaries
remains an under-explored area, especially for languages with complex linguistic
features like Turkish. This gap has the effect of impeding effective communication
and also limits the accessibility of information. Controlling readability of textual data
is an important element for creating summaries for different audiences with varying
literacy and education levels, such as students ranging from primary school to graduate
level, as well as individuals with diverse educational backgrounds. Summaries that
align with the needs of specific reader groups can improve comprehension and
engagement, ensuring that the intended message is effectively communicated. Furthermore, readability adjustment is essential to expand the usability of
summarization models in educational and professional domains.
Current summarization models often don’t have the mechanisms to adjust the
complexity of their outputs, resulting in summaries that may be too simplistic or overly
complex for certain types of reader groups. Developing adaptive models that can tailor
content to specific readability levels is therefore crucial. To address this problem, we create our own custom dataset and train a model with our custom architecture. Our method ensures that readability levels are effectively controlled while maintaining accuracy and coherence. We rigorously compare our model to a supervised fine-tuned baseline, demonstrating its superiority in generating readability-aware summaries. \newline

\end{abstract}

\newpage
\section{Introduction}
In recent years, natural language processing (NLP) has witnessed remarkable
progress, with the development of transformer-based models. These models not only significantly advanced the field of NLP, but also set new records in a wide range of NLP tasks, including summarization. The transformer model can process entire sequences at once, unlike recurrent models, which process sequences one step at a time. This parallelism enables transformers to capture long-range dependencies and complex linguistic patterns more efficiently and accurately \cite{vaswani2017attention}. As a result, transformer-based architectures have become the cornerstone of many NLP tasks, including text summarization.

However, the challenge of tailoring summaries to meet readability requirements remains largely unaddressed, despite these advances. Summarization without readability fails to meet the diverse needs of readers with varying literacy and educational backgrounds. Readability is a critical element to ensure effective
communication and requires the adjustment of linguistic complexity to meet the target audiences. However, current summarization models often lack the mechanisms to control readability for different groups of readers.

Readability control in text generation has recently gained attention with emerging techniques for conditional text generation offering promising solutions.Methods such as prompt
engineering \cite{prompteng}, fine-tuning with attribute-specific data \cite{ctrlpaper}, reinforcement learning \cite{ftfromhumanpref}, and steering \cite{steering} have demonstrated that large language models (LLMs) can generate outputs tailored to specific attributes like tone, style, and length. These approaches underscore the potential of LLMs to adapt their outputs to meet specific requirements, including readability levels. Integrating readability control into summarization models poses unique challenges, particularly for languages like Turkish. These challenges include developing suitable readability metrics that account for the complexities of Turkish morphology and syntax, as well as addressing the limited availability of annotated datasets for training and evaluation.

In this study, we take a focused approach to developing a readability-aware summarization model for Turkish. We leverage the power of transformer-based LLMs and integrate a readability metric specifically designed for the Turkish language. By embedding this customized metric into the summarization process, we ensure that our model generates summaries tailored to the comprehension levels of diverse target audiences.

\section{Related Works}
\label{sec:headings}

This chapter explores the development and evolution of summarization methods, starting with early extractive approaches that pick key sentences from the content to be summarized, progressing to abstractive techniques that generate new
sentences to capture the essence of the content, moving on to the discussion of readability metrics that evaluate the clarity and accessibility of the summaries, and concluding with strategies for controlling readability using large language models.

\subsection{Extractive Summarization}
Early approaches to automatic text summarization were mostly extractive, where the main idea is dependent on selecting the most important sentences from the source text to create a summary. These methods relied mainly on statistical techniques
to identify key information in textual data.

One of the first methods for summarization was proposed by \cite{luhn}, where the author used word frequency to select important sentences from a document. According to this approach, the importance of a sentence can be measured by counting the frequency of certain content words, or keywords, it contains.
The sentences are then given scores according to that frequency and ranked. The highest ranking scores are extracted to generate the summary.

\cite{edmunson} built upon this approach by utilizing additional features such as cue words, title and heading words, and sentence location. The method assigned weights to these features to score with linear summation and select sentences to create the summary.

Graph-based Models LexRank \cite{lexrank} and TextRank \cite{textrank} are foundational approaches that conceptualize documents as graphs and frame extractive summarization as the task of identifying the most central nodes within the graph, drawing inspiration from the PageRank algorithm \cite{pagerank}.

Machine learning techniques have taken the idea of extractive summarization one step further. Machine learning approaches model the summarization as a classification problem. \cite{kupiec} developed a trainable summarizer using a Bayesian classifier to determine the relevance of sentences. In this work, the authors created a naive Bayes classifier which differentiated between sentences to be included in a summary and those not to be included. This classifier uses features of the sentences
and is trained on a dataset comprising documents and their respective extractive summaries. Following the popularization of deep learning approaches, neural network models have been applied to extractive summarization tasks. \cite{chenlap} introduced a neural model that employs a hierarchical encoder to capture document structure and a sentence extractor to select important sentences.

Recent advances in transformer-based models have significantly improved extractive summarization. \cite{liutextsum} proposed a BERT-based extractive summarization model that achieves state-of-the-art performance by leveraging pre-trained language representations. Their approach demonstrates the effectiveness of
fine-tuning pre-trained models for specific summarization tasks.

\cite{discobert} introduced DISCOBERT, a discourse-aware neural extractive summarization model that operates at the sub-sentence level using Elementary Discourse Units (EDUs) instead of full sentences. It makes use of two types of discourse graphs: an RST graph and a co-reference graph. These graphs are capable of capturing long-range dependencies between units of text. Encoding these graphs using graph convolutional networks and combining them with representations from BERT allowed DISCOBERT to achieve state-of-the-art results. It proves that finer-grained
units for extraction using discourse structures can bring out much conciseness,  minimal redundancy, for informative summaries.

\subsection{Abstractive Summarization}
Abstractive summarization addresses the summarization problem by generating concise summaries by paraphrasing and condensing the source text, by adding new phrases and sentences not present in the original content. This approach requires a deep
understanding of the source content and advanced natural language generation capabilities and is akin to how people summarize texts. 

The advent of neural network models greatly impacted abstractive
summarization. An abstractive sentence summarization neural attention model was proposed by \cite{rush}. Their model was an encoder-decoder architecture coupled with attention mechanisms. The model successfully depicted the potential of
neural networks to generate coherent and relevant summaries. \cite{chopra} improved upon this work by incorporating a recurrent neural network (RNN) decoder, enhancing the model's ability to capture temporal dependencies in text generation.

\cite{nallap} explored the use of sequence-to-sequence RNNs for
abstractive summarization, introducing techniques to handle out-of-vocabulary words and capture document context. They employed a hierarchical encoder to process documents and a decoder with attention mechanisms to generate summaries.

\cite{seepointer} proposed a pointer-generator network, which combines the benefits of extractive and abstractive methods. This model allows for copying words directly from the source text (pointer mechanism), while retaining the ability to generate novel words, addressing issues like factual inaccuracies and handling rare words.

The introduction of the transformer architecture by \cite{vaswani2017attention} revolutionized natural language processing, including summarization tasks. Transformers rely entirely on attention mechanisms, enabling models to capture global dependencies more effectively than RNNs.

Pre-trained language models have further advanced abstractive summarization. \cite{liutextsum} introduced BERTSUM, which adapts BERT (Transformator Bidirectional Encoder Representations) to summarization tasks. BERTSUM extends BERT's architecture for both extractive and abstractive summarization by adding transformer layers for decoding.

\cite{lewisbart} presented BART, a denoising autoencoder that trains a sequence-to-sequence model by corrupting text and training the model to reconstruct the original. BART has shown strong performance on various summarization benchmarks, leveraging the strengths of both bidirectional and auto-regressive
transformers.

\cite{pegasus} introduced PEGASUS, a transformer-based model trained specifically for abstractive summarization. PEGASUS uses a novel training objective called Gap Sentences Generation (GSG), where important sentences are masked and the model learns to generate them from the remaining text. This approach closely simulates summarization tasks during training, resulting in improved performance.

Recent developments have focused on improving the factual consistency of abstractive summaries. \cite{dongcorrect} proposed a fact-aware summarization model that explicitly considers factual consistency during generation. Their approach
significantly reduces the occurrence of hallucinated content in generated summaries.

\cite{dousum} proposes a training objective, incorporating a guided
summarization framework that enhances semantic alignment between source
documents and generated summaries through auxiliary signals which maintains
factuality while still allowing for abstractive creativity.

\subsection{Readability Metrics}
Readability metrics are quantitative scores that help evaluate the ease or difficulty of reading a text. These metrics are derived from a range of linguistic features, including sentence length, word complexity, and syllable count. By analyzing
these metrics, insights can be gained regarding the clarity of a text. For languages like English and Turkish, several well-known readability formulas have been developed to measure written content.

\subsubsection{Flesch reading ease (FRES)}
A readability metric that evaluates how easy a
text is to understand, with higher scores indicating simpler language and easier comprehension, typically based on sentence length and syllable count per word \cite{fres}.
\begin{equation}
    FRES = (206.835 - (1.015 \times \text{ASL}) - (84.6 \times \text{ASW}))
\end{equation}

\noindent\text{where}
\begin{align*}
    &\text{ASL: Average sentence length.} \\
    &\text{ASW: Average number of syllables per word.}
\end{align*}

The formula for the Flesch reading ease score (FRES) calculates the readability of a text by considering the average sentence length (total words divided by total sentences)
and the average number of syllables per word (total syllables divided by total words), with coefficients that weight these factors to produce a score on a scale indicating text complexity.
\begin{table}[H]
\centering
\caption{Flesch reading ease (FRES) score interpretation.}
\vspace{10pt}
\label{tab:fres-interpretation}
\begin{tabular}{|c|c|}
\hline
\rowcolor{gray!20} Score & School level (US) \\
\hline
100.00--90.00 & 5th grade \\
\hline
90.0--80.0 & 6th grade \\
\hline
80.0--70.0 & 7th grade \\
\hline
70.0--60.0 & 8th \& 9th grade \\
\hline
60.0--50.0 & 10th to 12th grade \\
\hline
50.0--30.0 & College \\
\hline
30.0--10.0 & College graduate \\
\hline
10.0--0.0 & Professional \\
\hline
\end{tabular}
\end{table}

Table \ref{tab:fres-interpretation} shows FRES categories, mapping specific score
ranges to US school levels or stages of education. The higher the score, the easier the text is and the lower the score, the more complicated it is.

\subsubsection{Gunning fog index}
Gunning fog index (GFI) estimates the years of formal education needed to comprehend the text on the first reading. It considers sentence length and the percentage of complex words \cite{gunning}.
\begin{equation}
    \text{GFI} = 0.4 \times (\text{ASL} + 100 \times \text{PHW})
\end{equation}

\noindent\text{where}
\begin{align*}
    &\text{ASL: Average sentence length.} \\
    &\text{PHW: Percentage of hard words (any word with 3+ syllables,} \\
    &\quad\quad\quad\text{except words made complex by suffixes -ed or -es).}
\end{align*}
\begin{table}[H]
\centering
\caption{Gunning fog index.}
\vspace{10pt}
\label{tab:fog-index}
\begin{tabular}{|c|c|}
\hline
\rowcolor{gray!20} Fog Index & Reading level by grade \\
\hline
17 & College graduate \\
\hline
16 & College senior \\
\hline
15 & College junior \\
\hline
14 & College sophomore \\
\hline
13 & College freshman \\
\hline
12 & High school senior \\
\hline
11 & High school junior \\
\hline
10 & High school sophomore \\
\hline
9 & High school freshman \\
\hline
8 & Eighth grade \\
\hline
7 & Seventh grade \\
\hline
6 & Sixth grade \\
\hline
\end{tabular}
\end{table}
Table \ref{tab:fog-index} shows Gunning fog index values mapped to school grade levels. Higher values indicate more complex texts, while lower values reflect simpler and easier-to-read
texts.

\subsubsection{SMOG Index}
SMOG index estimates the years of education required to understand a piece of writing. It focuses on the number of polysyllabic words per sentence \cite{smog}.
\begin{equation}
    \text{SMOG} = 1.0430 \times \sqrt{\text{PC} \times \frac{30}{\text{SC}}} + 3.1291
\end{equation}
%diğer tabloların açıklamalarının sonuna nokta eklemişsiniz hocam ama bu eksik kalmış. Eklenmeli mi?
\noindent\text{where}
\begin{align*}
    &\text{PC: Polysyllable count.} \\
    &\text{SC: Total number of sentences.}
\end{align*}
\begin{table}[H]
\centering
\caption{SMOG Index}
\vspace{10pt}
\label{tab:smog-index}
\begin{tabular}{|c|c|}
\hline
\rowcolor{gray!20} Score/Grade & Education Level \\
\hline
1-4 & Elementary School \\
\hline
5-8 & Middle School \\
\hline
9-12 & High School \\
\hline
13-16 & Undergraduate \\
\hline
17+ & Graduate \\
\hline
\end{tabular}
\end{table}
Table \ref{tab:smog-index} maps SMOG index scores to education levels. Higher scores reflect texts with more complex vocabulary that require greater reading proficiency, while lower scores indicate simpler texts.

\subsubsection{ Automated readability index (ARI)}
ARI index computes the US grade level needed to comprehend the text, using characters per word and words per sentence \cite{ari}.
\begin{equation}
    \text{ARI} = 4.71 \times \text{AWL} + 0.5 \times \text{ASL} - 21.43
\end{equation}

\noindent\text{where}
\begin{align*}
    &\text{AWL: Average number of characters per word.} \\
    &\text{ASL: Average number of words per sentence.}
\end{align*}
\begin{table}[H]
\centering
\caption{Automated readability index.}
\vspace{10pt}
\label{tab:ari}
\begin{tabular}{|c|c|c|}
\hline
\rowcolor{gray!20} Score & Age & Grade Level \\
\hline
1 & 5-6 & Kindergarten \\
\hline
2 & 6-7 & First Grade \\
\hline
3 & 7-8 & Second Grade \\
\hline
4 & 8-9 & Third Grade \\
\hline
5 & 9-10 & Fourth Grade \\
\hline
6 & 10-11 & Fifth Grade \\
\hline
7 & 11-12 & Sixth Grade \\
\hline
8 & 12-13 & Seventh Grade \\
\hline
9 & 13-14 & Eighth Grade \\
\hline
10 & 14-15 & Ninth Grade \\
\hline
11 & 15-16 & Tenth Grade \\
\hline
12 & 16-17 & Eleventh Grade \\
\hline
13 & 17-18 & Twelfth Grade \\
\hline
14 & 18-22 & College Student \\
\hline
\end{tabular}
\end{table}
Table \ref{tab:ari} maps automated readability index (ARI) scores to US grade levels, with higher scores indicating more complex texts.

\subsubsection{Ateşman Readability Formula}
Ateşman readability formula is the first Turkish readability formula, adapted
from Flesch Reading Ease to account for the characteristics of the Turkish language. It
produces a score between 0-100, where higher scores indicate easier readability.\cite{ateşman}
\begin{equation}
    \text{Ateşman Readability Score} = 198.825 - 40.175 \times \text{ASW} - 2.610 \times \text{ASL}
\end{equation}

\noindent\text{where}
\begin{align*}
    &\text{ASW: Average number of syllables per word.} \\
    &\text{ASL: Average sentence length.}
\end{align*}
\begin{table}[H]
\centering
\caption{Ateşman readability index.}
\vspace{10pt}
\label{tab:atesman-scores}
\begin{tabular}{|c|c|}
\hline
\rowcolor{gray!20} Difficulty Level & Readability Score \\
\hline
Very Easy & 90-100 \\
\hline
Easy & 70-89 \\
\hline
Moderately Difficult & 50-69 \\
\hline
Difficult & 30-49 \\
\hline
Very Difficult & 1-29 \\
\hline
\end{tabular}
\end{table}
Table \ref{tab:atesman-scores} shows Ateşman readability scores, adapted for Turkish texts.
Scores range from 0 to 100, with higher scores indicating easier readability and lower
scores reflecting more complex texts.

\subsubsection{Çetinkaya-Uzun Readability Formula}
Developed in 2010, this formula was
created specifically for Turkish texts through scientific processes and validated using
cloze tests. It considers Turkish linguistic characteristics and provides educational
level classifications.\cite{cetinkaya}
\begin{equation}
    \text{RS} = 118,823 - (25,987 \times \text{AWL}) - (0,971 \times \text{ASL})
\end{equation}

\noindent\text{where}
\begin{align*}
    &\text{RS: Readability score.} \\
    &\text{AWL: Average word length.} \\
    &\text{ASL: Average sentence length.}
\end{align*}
\begin{table}[H]
\centering
\caption{Çetinkaya-Uzun readability index.}
\vspace{10pt}
\label{tab:cetinkaya-uzun}
\begin{tabular}{|c|c|c|}
\hline
\rowcolor{gray!20} Readability Score & Readability Level & Grade Level \\
\hline
0-34 & Insufficient Reading Level & 10th, 11th, and 12th Grade \\
\hline
35-50 & Educational Reading Level & 8th and 9th Grade \\
\hline
51+ & Independent Reading Level & 5th, 6th, and 7th Grade \\
\hline
\end{tabular}
\end{table}
Table \ref{tab:cetinkaya-uzun} shows Çetinkaya-Uzun readability scores mapped to readability
levels and grade classifications. Higher scores indicate texts that are easier to read,
while lower scores reflect more challenging texts that require advanced reading skills.

\subsubsection{Bezirci-Yılmaz Readability Formula}
\label{sec:yod_section}
Bezirci-Yılmaz readability formula defines the YOD readability metric specifically designed for Turkish texts. It calculates the readability
score based on the average number of polysyllabic words (three or more syllables)
per sentence. The metric assigns weights to these polysyllabic words and combines
them with the average sentence length, providing a clearer assessment of text
complexity \cite{yod}.
%clearer yerine clear olmalı mı?
\begin{equation}
    \text{YOD} = \sqrt{\text{OKS} \times ((H3 \times 0.84) + (H4 \times 1.5) + (H5 \times 3.5) + (H6 \times 26.25))}
\end{equation}
%burda hepsinin başına The yazmışım ama gerek yok gibi.
\noindent\text{where}
\begin{align*}
    &\text{OKS: The average number of words per sentence.} \\
    &\text{H3: The average number of three-syllable words per sentence.} \\
    &\text{H4: The average number of four-syllable words per sentence.} \\
    &\text{H5: The average number of five-syllable words per sentence.} \\
    &\text{H6: The average number of six or more syllable words per} \\
    &\quad\;\text{sentence.}
\end{align*}
\begin{table}[H]
\centering
\caption{Bezirci-Yılmaz readability index.}
\vspace{10pt}
\label{tab:bezirci-yilmaz}
\begin{tabular}{|c|c|}
\hline
\rowcolor{gray!20} YOD Value & Education Level \\
\hline
1--8 & Elementary School \\
\hline
9--12 & High School \\
\hline
13--15 & Undergraduate Level \\
\hline
16 and above & Academic/Professional Level \\
\hline
\end{tabular}
\end{table}
Table \ref{tab:bezirci-yilmaz} shows Bezirci-Yılmaz readability formula values mapped to educational levels. Higher YOD values indicate texts with greater complexity, requiring advanced
reading skills, while lower values correspond to texts that are easier to understand.

\subsection{Readability Control with Large Language Models}
Controlling the readability level of the generated summaries is crucial for tailoring the
content to specific audiences, such as language learners, children, or individuals with
varying literacy levels. \cite{scarton} investigated learning text
simplifications for specific target audiences using neural models. They proposed a
methodology to simplify text conditioned on the target audience's proficiency level,
enhancing accessibility.

\cite{martin} introduced ACCESS (AudienCe-CEntric Sentence Simplification), a controllable sequence-to-sequence model for text simplification that allows control over attributes such as simplicity, paraphrasing, and sentence length. By manipulating these attributes, the model can generate text at different readability levels. \cite{ctrlpaper} developed CTRL (Conditional Transformer Language
Model), which uses control codes to generate text with specific styles and attributes. By conditioning on these control codes, CTRL can produce text that adheres to desired
characteristics, including readability.

The emergence of large language models like GPT-3 \cite{prompteng} has enabled advanced controllable text generation with prompt engineering. With proper engineered prompts that include a needed style or level of reading, GPT-3 can generate the same level in kind. \cite{kikuchi} explored controlling the output length in neural encoder-decoders, which indirectly affects readability by influencing sentence complexity. Their method allows for the generation of summaries with specified
lengths, aiding in readability control.

\cite{ctrlsum} presented CTRLSUM, a framework for controllable text summarization using keyword-driven approaches. Using keywords as control signals, CTRLSUM allows the generation of summaries tailored to specific
dimensions, including readability, length, and entity focus. \cite{ribeiro}
introduced methods for generating summaries at specified readability levels. Their
approach integrates instruction-based control, reinforcement learning, and lookahead
decoding, enabling fine-grained readability adjustment to align with target
audiences. \cite{laam} proposed a Length-Aware Attention Mechanism (LAAM)
to produce summaries with precise length constraints. This method adapts the
encoding and attention mechanism to align with desired lengths, indirectly influencing
readability by tailoring content scope. \cite{krishna} explored controllable style
transfer using style vectors derived from paraphrases, which enables fine control over
stylistic attributes, particularly in low-resource multilingual settings. \cite{luo}
introduced a framework for readability-controllable biomedical document
summarization. By generating both technical summaries for experts and plain-language summaries for lay readers, their approach improves accessibility while
addressing diverse audience needs.

\section{Methodology}
\subsection{Large Language Model}
In Turkish NLP, there is a significant gap in the availability of large language
models (LLMs) tailored specifically for the language. Turkish is often
underrepresented in global NLP efforts, as most LLMs are either multilingual,
prioritizing high-resource languages, or are too large to be fine-tuned effectively due
to resource constraints. Recognizing this challenge, VNGRS addressed the issue by
developing VBART\cite{vbart}, a family of Turkish encoder-decoder sequence-to-sequence
LLMs, designed specifically to advance NLP tasks in Turkish.

For text summarization, VBART comprises two additional variants: VBARTLarge (387 million parameters) and VBART-XLarge (744 million parameters). These
models build on the mBART architecture and feature enhancements such as sinusoidal
positional embeddings, which improve training stability in mixed precision settings.
Unlike multilingual models, VBART focuses entirely on Turkish, offering significant
advantages in efficiency and performance, especially for low-resource tasks.

In this study, the VBART-Large model was used, with VBART-Large chosen for its strong balance between performance and efficiency. Although VBART-XLarge delivers better results, its double size compared to VBART-Large makes the marginal 
performance gains impractical for this study.
%Yukarıdaki additionally, the other two ... kısmına ben görsel eklemiştim ordan refere edip söylüyorum ama görseli kaldırdığımız için bu cümleye gerek yok gibi.
\subsection{Readibility Metric}
This study employs the YOD\footnote{In Turkish, YOD stands for Yeni Okunabilirlik Düzeyi.} formula, selected for
its ability to capture the unique linguistic features of Turkish, such as its agglutinative
structure and variability in word length. Unlike traditional English-based metrics,
YOD provides a more relevant and accurate measure of the readability of Turkish text, which makes it ideal for this study.

The YOD formula was designed to assess the readability of Turkish texts, addressing the specific linguistic features of the language
that differ significantly from English. Traditional readability formulas such as SMOG,
Gunning-Fog, and Flesch are based on the structural characteristics of English and do
not account for the unique morphological and syntactic complexities of Turkish. As
an agglutinative language, Turkish forms words by adding multiple suffixes to root
words, creating substantial variation in word length and structure. This required a
specialized approach tailored to Turkish.

To develop the formula, the researchers analyzed 20 different types of books
and magazines. They calculated key linguistic metrics such as average word length,
sentence length, and frequency of words with varying syllable counts. The analysis
revealed that one-syllable and two-syllable words dominated all texts, regardless of
complexity or genre. These high-frequency words, like "bu" (this), "şu" (that), and
"ve" (and), were excluded from the formula as they contribute minimally to text
complexity. In contrast, words with three, four, five, six, or more syllables showed
significant variation across genres, making them reliable indicators of textual
difficulty. This observation underscored the importance of focusing on multi-syllabic
words in assessing readability. The readability of Turkish texts is assessed using the YOD formula, which can be found in Sec. \ref{sec:yod_section}.

\subsection{Dataset}
Although there is a scarcity of Turkish summarization datasets in comparison
to those available for high-resource languages such as English, several datasets
have emerged in recent years to support Turkish natural language processing tasks\cite{baykara} \cite{musabg2023wikipediatrsummarization} \cite{xlsum} \cite{mlsum} \cite{lrsum}.
Despite their limited number and size, these datasets are of value for the development
and evaluation of Turkish language summarization models. The experimental dataset
for this study was constructed by sampling from five major Turkish summarization
datasets, with selection controlled by the model’s maximum token limit. Sampling is
done hierarchically from longest text that can fit into tokenizer without truncation to the shortest content.

\begin{table}[H]
\centering
\caption{Dataset splits for train, validation, and test.}
\vspace{10pt}
\label{tab:dataset-splits}
\begin{tabular}{|c|c|c|c|}
\hline
\rowcolor{gray!20} Dataset & Train & Validation & Test \\
\hline
TR-News & 277,573 & 14,610 & 15,379 \\
\hline
wikipedia-tr-summarization & 119,110 & 6,269 & - \\
\hline
XLSum & 27,176 & 3,397 & 3,397 \\
\hline
MLSUM & 249,277 & 11,565 & 12,775 \\
\hline
LR-Sum & 28,672 & 3,583 & 3,584 \\
\hline
\end{tabular}
\end{table}

The available Turkish summarization datasets provided in Table \ref{tab:dataset-splits}, while valuable for developing
and evaluating language models, exhibit a notable shortcoming in terms of readability awareness when assessed with the YOD formula. This limitation prevents the effective
fine-tuning of large language models that aim to generate summaries with controlled
readability levels.

\begin{table}[H]
\centering
\caption{Distribution of abstracts in the train split assessed using the YOD formula.}
\vspace{10pt}
\label{tab:yod-distribution}
\scalebox{0.90}{
\begin{tabular}{|c|c|}
\hline
\rowcolor{gray!20} YOD & Count \\
\hline
1 & 2,543 \\
\hline
2 & 54,092 \\
\hline
3 & 164,786 \\
\hline
4 & 156,022 \\
\hline
5 & 80,296 \\
\hline
6 & 108,333 \\
\hline
7 & 60,130 \\
\hline
8 & 39,298 \\
\hline
9 & 19,246 \\
\hline
10 & 10,279 \\
\hline
11 & 4,124 \\
\hline
12 & 1,693 \\
\hline
13 & 637 \\
\hline
14 & 215 \\
\hline
15 & 76 \\
\hline
16 & 38 \\
\hline
\end{tabular}
}
\end{table}
Table \ref{tab:yod-distribution} illustrates the distribution of YOD values across the training sets, showing
a clear concentration of abstracts at lower readability levels. Higher YOD values, while present, appear with significantly lower frequencies.
%the VBART-Large-Paraphrasing model burda the gerekli değil gibi
%% Bunu çok anlamadım ama kalabilir gibi duruyor.
To address this gap, the VBART-Large-Paraphrasing model was employed to enhance the existing datasets by generating paraphrased variations at both the sentence
and full-text levels. This approach permitted the derivation of content with a more extensive range of YOD values, encompassing both higher and lower values, from the
same source material. To maintain semantic integrity, each paraphrase was compared to the original summary using BERTScore to verify that the synthetic data achieved
the intended readability adjustments while remaining faithful to the source. In addition, ChatGPT’s API was also used for synthetic data generation, enriching
the dataset with diverse and high-quality rewritten summaries. 
\begin{figure}[H]
    \centering
    \includegraphics[width=1\linewidth]{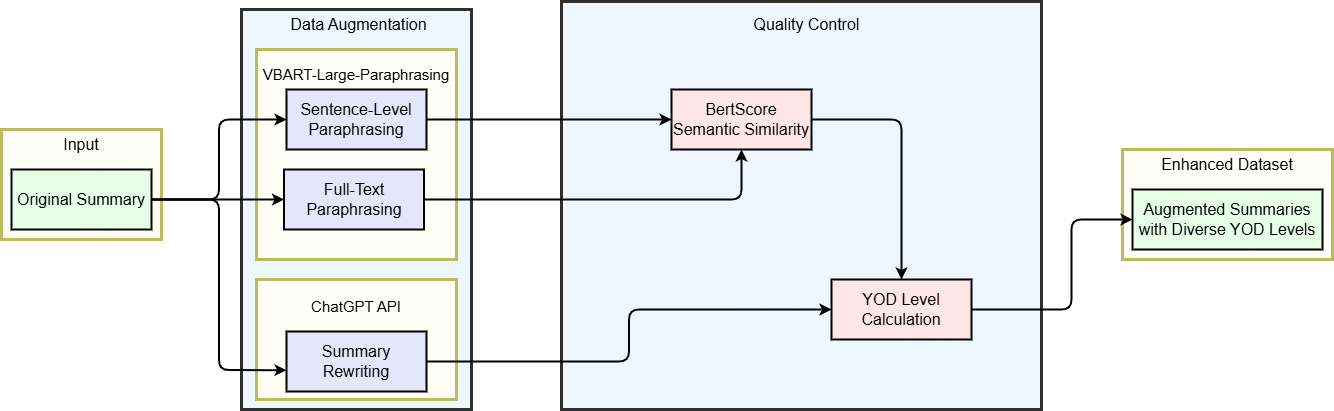}
    \caption{Synthetic Data Generation}
    \label{fig:data_augmentation}
\end{figure}

The strategy shown in \ref{fig:data_augmentation} facilitated the creation of a dataset which contains more balanced YOD values. Although the YOD formula allows for readability scores above 16, we have limited the scores up to 16 for the purposes of this study, thereby
ensuring consistency and alignment with our research objectives. Furthermore, according to the formula, there is no noticeable difference between scores of 16 and those above.

The new dataset is compiled from multiple sources: XLSUM (970 entries), TRNews (5,000 entries), MLSUM (1,033 entries), LR-SUM (1,107 entries), and
Wikipedia-tr-summarization (3,024 entries). After the synthetic data generation
process, the dataset is significantly expanded to include 76,759 summaries. To
guarantee a thorough evaluation, 200 samples for each YOD level are allocated to both
the test and validation sets, resulting in a total of 3200 examples for both test and
evaluation.

\begin{table}[H]
\centering
\caption{Synthetically generated dataset splits by YOD values.}
\vspace{10pt}
\label{tab:yod-splits}
\scalebox{0.80}{
\begin{tabular}{|c|c|c|c|c|}
\hline
\rowcolor{gray!20} YOD & Train & Test & Validation & Total \\
\hline
1 & 1,915 & 200 & 200 & 2,315 \\
\hline
2 & 7,147 & 200 & 200 & 7,547 \\
\hline
3 & 8,731 & 200 & 200 & 9,131 \\
\hline
4 & 7,184 & 200 & 200 & 7,584 \\
\hline
5 & 5,129 & 200 & 200 & 5,529 \\
\hline
6 & 4,182 & 200 & 200 & 4,582 \\
\hline
7 & 3,548 & 200 & 200 & 3,948 \\
\hline
8 & 3,862 & 200 & 200 & 4,262 \\
\hline
9 & 3,939 & 200 & 200 & 4,339 \\
\hline
10 & 3,816 & 200 & 200 & 4,216 \\
\hline
11 & 3,413 & 200 & 200 & 3,813 \\
\hline
12 & 2,863 & 200 & 200 & 3,263 \\
\hline
13 & 2,369 & 200 & 200 & 2,769 \\
\hline
14 & 1,872 & 200 & 200 & 2,272 \\
\hline
15 & 1,828 & 200 & 200 & 2,228 \\
\hline
16 & 8,561 & 200 & 200 & 8,961 \\
\hline
\rowcolor{gray!20} Total & 70,359 & 3,200 & 3,200 & 76,759 \\
\hline
\end{tabular}
}
\end{table}

Table \ref{tab:yod-splits} presents the new synthetically generated dataset, showing item counts across the train, test, and validation splits for each YOD value. The Total
column provides the overall number of items for each YOD level, demonstrating the
distribution and balance of data across different readability levels.

The dataset is systematically divided into three parts: train, test, and validation, covering YOD levels from 1 to 16. As YOD levels increase, text lengths increase steadily and cover different readability levels. In addition, the overall text lengths in all sections show a suitable distribution for
practical training and evaluation of the model. The assignment of fewer words for low YOD
values and the gradual increase in the number of words for higher YOD values
improves readability, bringing the dataset in line with the complexity and progression of natural text.

\begin{figure}[H]
    \centering
    \includegraphics[width=1\linewidth]{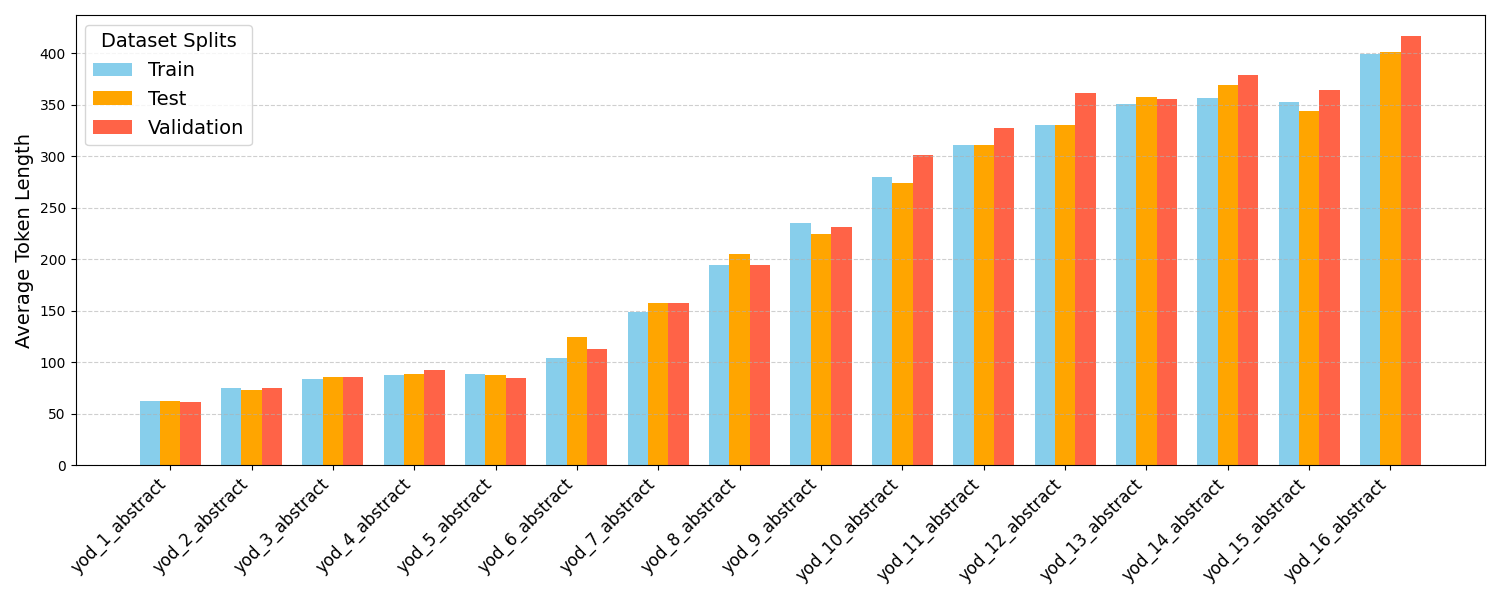}
    \caption{Average Token Length}
    \label{fig:token_length}
\end{figure}

Figure \ref{fig:token_length} highlights a clear upward trend, where higher YOD positions correspond
to longer token counts, reflecting increased text complexity at higher readability levels.

\subsection{Custom Model Architecture}
This part presents our multi-task architecture for Turkish text summarization
based on VBART (provided by VNGRS-AI), integrated with readability level (YOD
- Yeni Okunabilirlik Düzeyi) prediction. This model, which is fundamentally built on
an encoder-decoder structure, generates summaries and, at the same time, predicts
readability scores and classes for input texts. In this way, not only is the summary
created, but also the controllability of the generated summary is enhanced by providing
additional information aligned with the desired readability level.

\begin{figure}[H]
    \centering
    \includegraphics[width=1\linewidth]{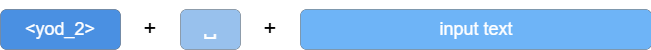}
    \caption{Special Token Conditioning}
    \label{fig:token_Conditioning}
\end{figure}

To incorporate readability levels into the model, special tokens ranging from
<yod\_1> to <yod\_16> are introduced, with each token corresponding to a specific
YOD value in the formula. These tokens serve as markers for the model in
understanding and generating text at the desired readability level. And the training process functions as follows: the model processes input texts, including special
<yod\_i> tokens (where i ranges from 1 to 16), which are placed at the beginning of the
input, as shown in \ref{fig:token_Conditioning} to specify the desired readability level. These tokens guide the model to interpret
or generate text aligned with the indicated YOD level. During tokenization, these
<yod\_i> tokens are integrated into the model, ensuring that the input is processed
according to the corresponding YOD level by interpreting these special symbols. This
approach improves the ability of the model to process texts according to their readability
requirements while maintaining contextual and linguistic coherence.

Following the encoding process, the resulting token-based representations are condensed into a fixed-size vector through mean pooling, facilitated by the attention mask. This pooled vector reflects the overall content of the input text and serves as a compact "semantic representation" for the subsequent YOD prediction tasks. The model incorporates two specialized sub-networks (prediction heads) that utilize this pooled representation.

\subsubsection{YOD Regression Head (YOD Regressor)}
This sub-network predicts a continuous YOD score by passing the encoder's pooled representation through a 4-layer MLP (encoder’s last hidden state → 512 → 256 → 128 → 1). It employs ReLU activations, dropout layers, and Xavier weight initialization to ensure effective training while minimizing overfitting. This head provides a detailed, numerical assessment of text readability.

\subsubsection{YOD Classification Head (YOD Classifier)}
This sub-network predicts discrete readability categories (from 1 to 16) using a similar MLP architecture (encoder’s last hidden state → 512 → 256 → 128 → 16). The resulting logits are mapped to YOD classes, which enables the model to provide a comprehensive perspective on text readability, complementing the continuous score with discrete categorization.\newline

These two sub-networks share the same pooled input representation (encoder mean), ensuring coherence between the continuous (regression) and categorical (classification) readability predictions. This dual approach enriches the model's ability to assess readability from multiple perspectives. The training process optimizes three distinct loss functions simultaneously. This design not only allows the model to generate summaries, but also provides controlled readability predictions, enabling better control and adaptability in output generation. By integrating both regression and classification approaches for YOD prediction, the model delivers robust and versatile readability assessments, tailored to the specified requirements.
\begin{figure}[H]
    \centering
    \includegraphics[width=1\linewidth]{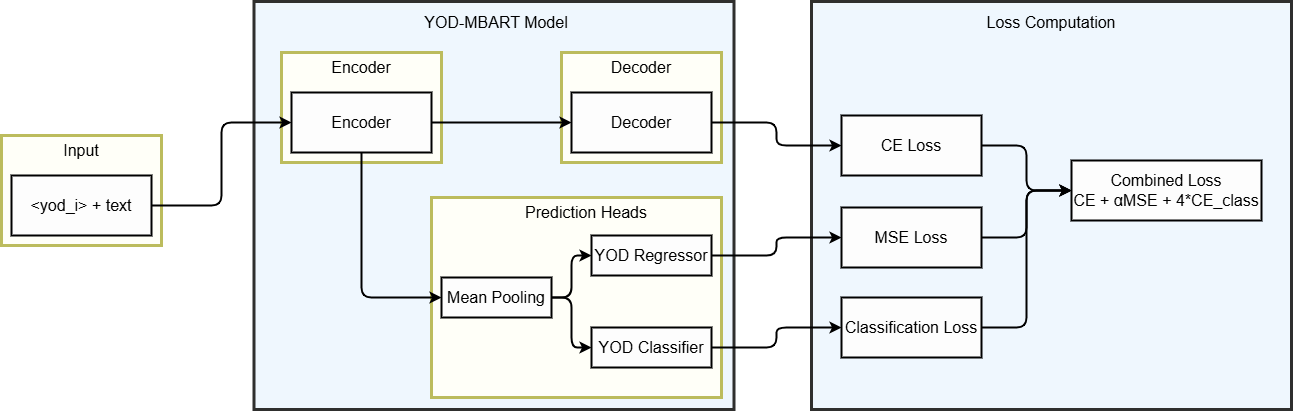}
    \caption{Custom Model Architecture
}
    \label{fig:enter-label}
\end{figure}
\subsection{Training}
The training methodology for the proposed architecture implements a multi-objective optimization framework that balances three core tasks: abstractive summarization, continuous YOD prediction, and YOD classification. This process employs modern training techniques, dynamic loss weighting, and hardware optimizations to ensure both efficient and effective training. Additionally, a weighted sampling strategy is utilized to address class imbalances across YOD categories, ensuring equal representation and learning for all readability levels during training.

The model is trained using the AdamW optimizer with a fused implementation to improve computational efficiency. A cosine learning rate schedule is employed, with an initial learning rate of  $1 \times 10^{-5}$ and a warmup of 500 steps to stabilize the training process. The training spans 11 epochs, with a batch size of 64 per device, enabling effective utilization of hardware resources. Mixed precision training (FP16) and gradient check-pointing are used to reduce memory usage and improve computational efficiency. Additional configurations include weight decay (0.01) to prevent overfitting and gradient clipping (1.0) to manage gradient stability.

\subsection{Loss Function Architecture}
The model’s loss function is composed of three components, each serving a distinct purpose in guiding the training process toward readability-controlled summarization:

\subsubsection{Cross-Entropy Loss (CE Loss)}
This loss ensures that the model generates summaries closely aligned with the provided target summaries. By comparing the predicted token distributions to the ground-truth tokens, the CE loss encourages the model to produce contextually relevant and accurate summaries.

\subsubsection{Mean Squared Error (MSE) Loss for YOD Regression}
This regression-based objective measures how well the model predicts a continuous YOD readability score. Minimizing this loss encourages the model to produce summaries whose difficulty level closely matches a specified numerical target, allowing for finer-grained readability adjustments.

\subsubsection{Cross-Entropy Loss for YOD Classification}
In addition to predicting continuous readability levels, the model also categorizes summaries into discrete readability classes. The classification loss guides the model to identify the correct YOD class, enabling a more interpretable and controllable readability output.

The total loss is formulated as in Eq. \ref{eq:loss}
\begin{equation}
\label{eq:loss}
    \mathcal{L}_{\text{total}} = \mathcal{L}_{\text{CE}} + w_{\text{YOD}}\mathcal{L}_{\text{YOD}} + 4 \cdot \mathcal{L}_{\text{class}}
\end{equation}
\noindent\text{where}
\begin{align*}
    &w_{\text{YOD}}: \text{dynamic weight, starting at 0.4 and increasing by 0.05 every 3 epochs (max 0.8)} \\
   &4: \text{fixed classification weight to maintain task importance during training} \\
&\quad \text{(This weight is experimentally found to give the best results)}
\end{align*}

This dynamic weighting strategy allows the model to initially prioritize summarization, gradually increasing emphasis on readability prediction tasks as training progresses.

\subsection{Training Results}
For a fair evaluation and comparison, the same model was fine-tuned with supervised tuning, trained for 11 epochs, and incorporated class imbalance handling, as implemented in the custom-architectured model. The performance of the fine-tuned model was then compared to that of the custom-architectured model, with a detailed analysis conducted to highlight the strengths and weaknesses of each approach. The evaluation of each summary was based on two primary criteria:

\begin{enumerate}
    \item Semantic similarity to the original text, assessed using standard metrics such as\cite{rouge} \cite{meteor} \cite{bleu}: 
    \begin{itemize}
        \item ROUGE-1 (unigram overlap)
        \item ROUGE-2 (bigram overlap)
        \item ROUGE-L (longest common subsequence)
        \item METEOR (explicit word-to-word matches)
        \item BLEU (precision-focused evaluation)
    \end{itemize}
    
    \item The ability of the summary to achieve the target reading level within the specified YOD tolerance range. The $\pm1.5$ tolerance in YOD evaluation is added specifically for neural language model assessment. When AI models generate text, they make word-by-word decisions based on probabilities, making it challenging to hit exact readability score targets. This added tolerance acknowledges that small variations in text features (like sentence length and word complexity) are normal and expected in AI-generated content.
\end{enumerate}

The comparisons below provide detailed insights into each model's performance across ROUGE metrics, BLEU scores, and YOD assessments, demonstrating their effectiveness in both maintaining semantic accuracy and achieving target readability levels.
\begin{figure}[H]
    \centering
    \includegraphics[width=1\linewidth]{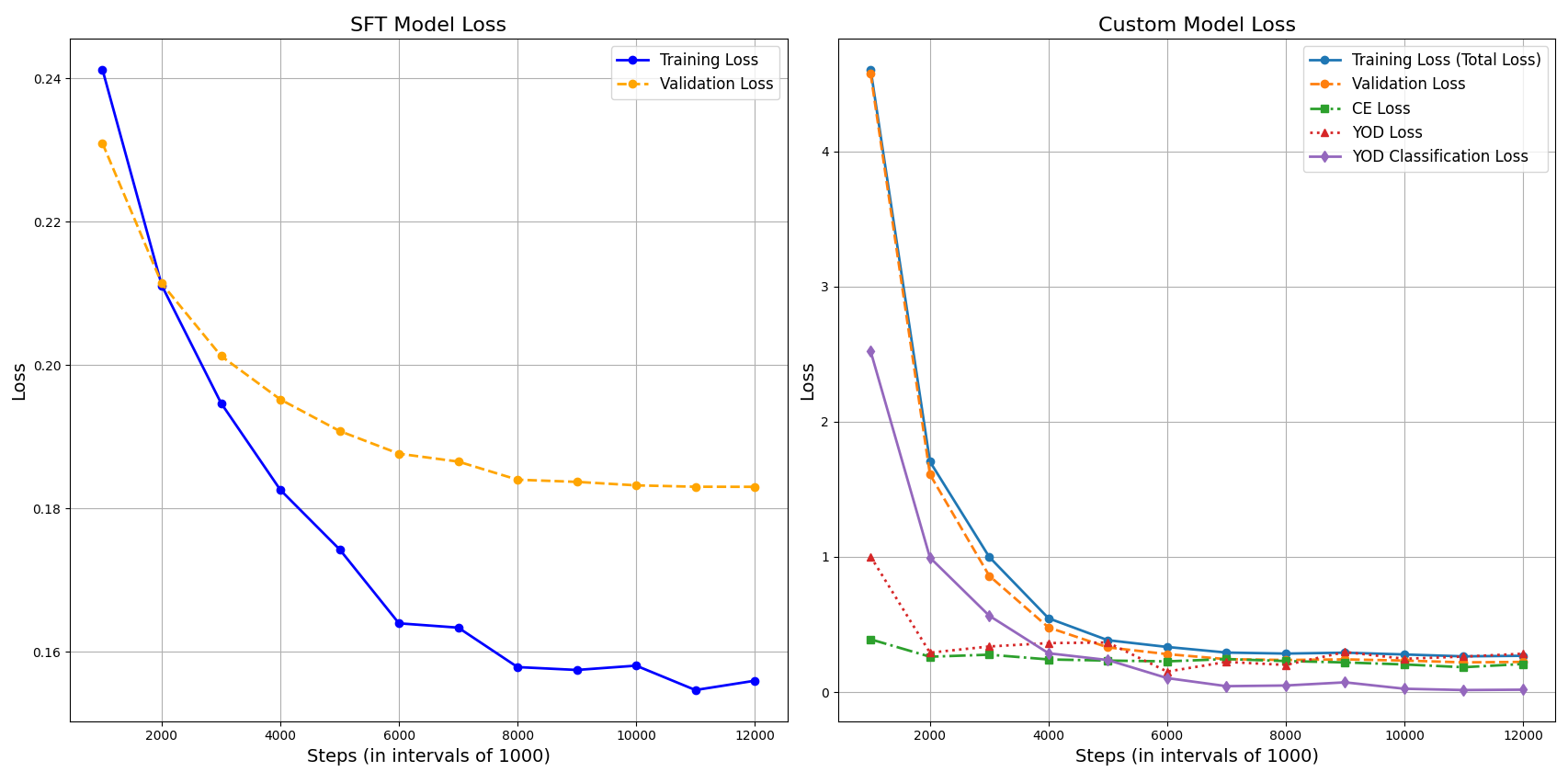}
    \caption{Loss curves comparison – supervised fine-tuned model vs. custom architecture.}
    \label{fig:loss_comparison}
\end{figure}

Figure \ref{fig:loss_comparison} shows the loss curves of the SFT (Supervised Fine-Tuning) model (on the left) and the customized model (on the right). The customized model has a more complex learning process as it trains three different neural networks simultaneously and therefore has higher initial losses. 

The learning process of the SFT model shows a consistent and gradual improvement over time. Starting from a low initial value of 0.24, the training loss decreases steadily with every 1000 steps, reaching 0.156 by the 12,000th step. Similarly, the validation loss is initially 0.23 and gradually decreases over time, stabilizing at 0.183. 

On the other hand, the customized model demonstrates its ability to perform complex learning tasks effectively, despite starting with higher initial losses. The training loss initially started at a high value of 4.60, but decreased rapidly to 1.70 in the first 1000 steps. A similar pattern was observed in the validation loss, which rapidly decreased from 4.57 to 1.60. Between steps 1,000 and 3,000, the losses show a consistent decrease, reaching a minimum of 0.54 for the training loss and 0.47 for the validation loss. Between steps 3,000 and 5,000, the improvements are more gradual, with the training loss reducing to 0.38 and the validation loss to 0.33. From step 5,000 onward, both losses stabilize, with the training loss at approximately 0.28 and the validation loss at around 0.23.

Despite an initial performance that was considerably less optimal, the customized model demonstrated notable advancements and exhibited enhanced functionality.  In conclusion, while the SFT model showed a more regular and smooth learning curve, the customized model stood out with a fast initial learning and then demonstrated its potential to successfully perform more complex tasks with a steady convergence process.

\begin{figure}[H]
    \centering
    \includegraphics[width=1\linewidth]{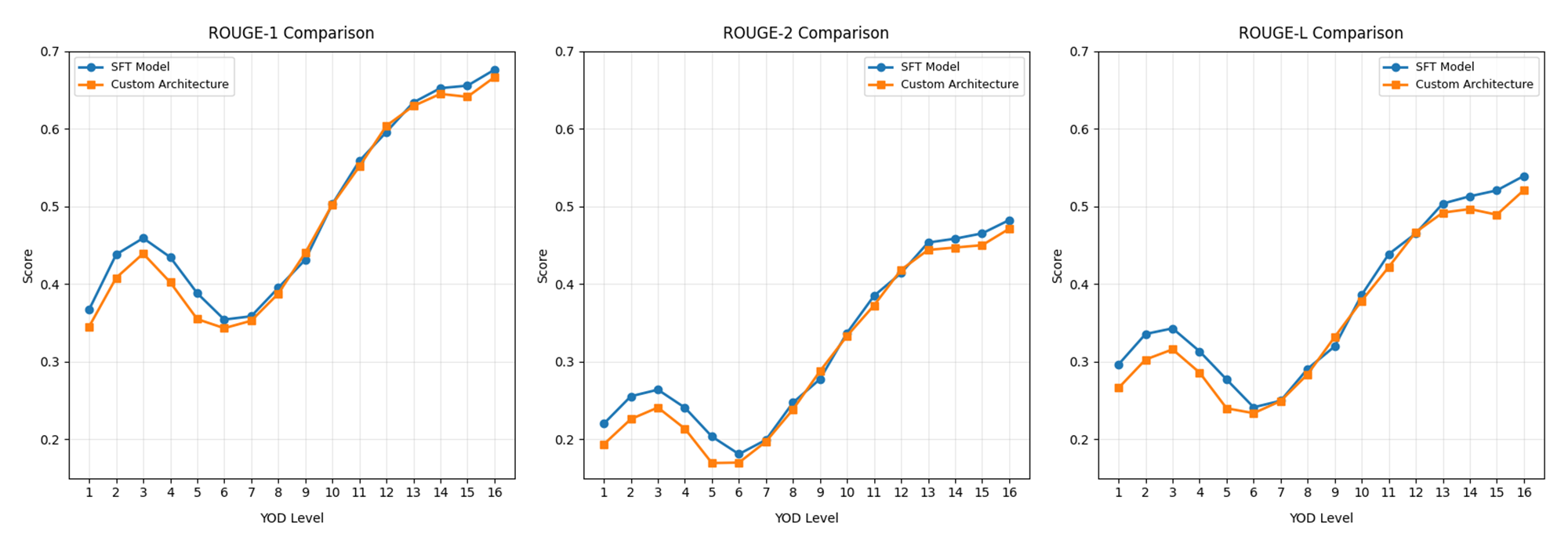}
    \caption{ROUGE score comparison – SFT model vs. custom architecture.}
    \label{fig:rouge_comp}
\end{figure}

Figure \ref{fig:rouge_comp} shows the comparison of the text summarization performance of SFT and custom-architectured models using ROUGE metrics at 16 YOD levels. The results reveal similar performance between the two approaches, with only minimal differences in effectiveness.

Both models exhibit strong unigram matching capabilities, as reflected in their ROUGE-1 scores. The SFT model progresses from 0.367 to 0.676 by YOD 16, while the custom architecture advances from 0.345 to 0.666. The final difference of just 0.009 demonstrates near-identical performance in capturing individual words. Similarly, bigram matching (ROUGE-2) shows comparable improvement, with the SFT model ranging from 0.220 to 0.482 and the custom architecture from 0.193 to 0.471. The final gap of 0.011 (about 1.1\%) reflects minimal differences in capturing word pairs. For sequence matching (ROUGE-L), the SFT model achieves scores from 0.296 to 0.539, while the custom architecture spans from 0.266 to 0.520. The largest difference of 0.018 (about 1.8\%) remains minor.

Both models show a similar trend, including a temporary dip around YOD levels 5-6, a significant improvement from YOD 9-13, and performance stabilization at higher YOD levels (13-16).

\begin{figure}[H]
    \centering
    \includegraphics[width=1\linewidth]{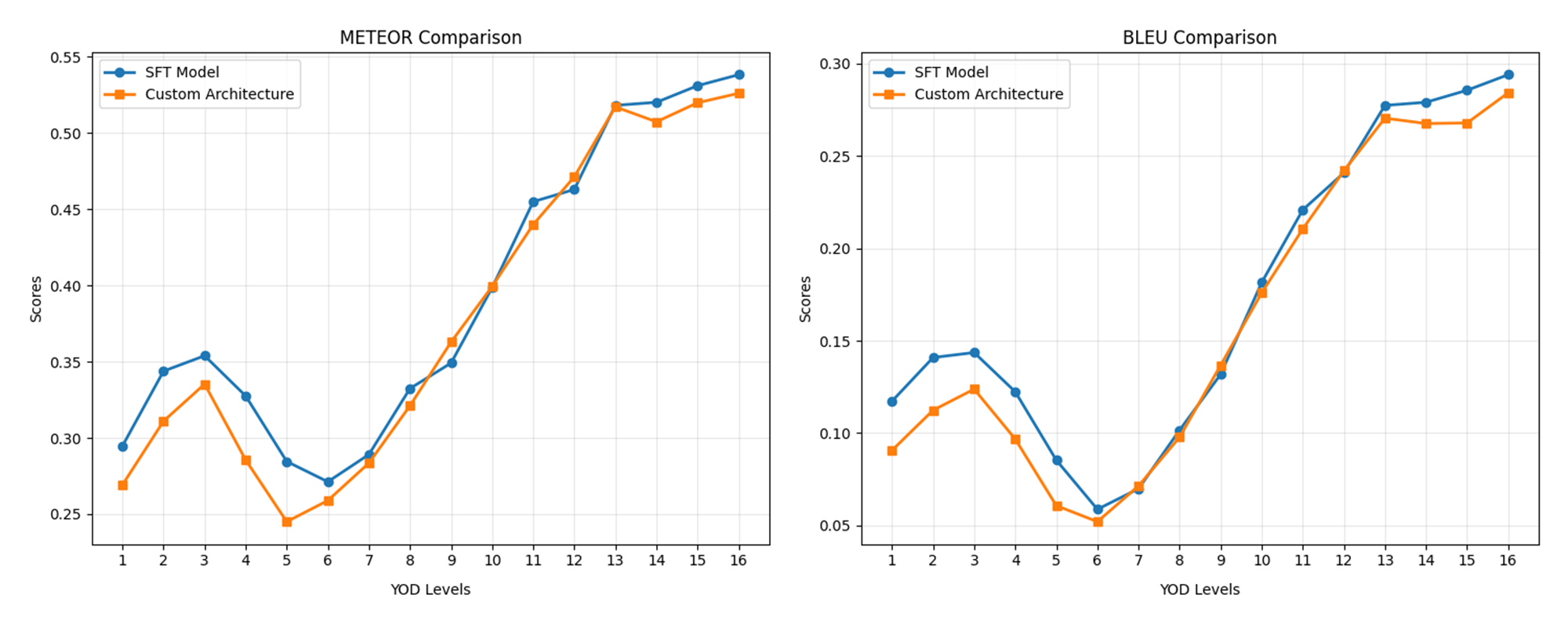}
    \caption{METEOR and BLEU score comparison – SFT Model vs. custom architecture.}
    \label{fig:blue_comp}
\end{figure}

Figure \ref{fig:blue_comp} shows the comparison of the text summarization performance of SFT and custom-architectured models using using METEOR and BLEU metrics across 16 YOD levels. The results indicate similar performance between the two approaches, with slight differences favoring the SFT model.

In METEOR, the SFT model progresses from 0.294 to 0.538, while the custom architecture ranges from 0.269 to 0.526. The final difference of 0.012 reflects a small edge for the SFT model in semantic alignment. Similarly, in BLEU, the SFT model scores range from 0.117 to 0.293, compared to 0.090 to 0.284 for the custom architecture, with a final gap of 0.009 indicating slightly better precision.

Both models follow similar trends, including a temporary dip around YOD levels 5-6, steady improvement from YOD 9-13, and stabilization at higher levels. These results suggest that both models are effective, with differences that are unlikely to significantly impact practical applications.

\begin{figure}[H]
    \centering
    \includegraphics[width=1\linewidth]{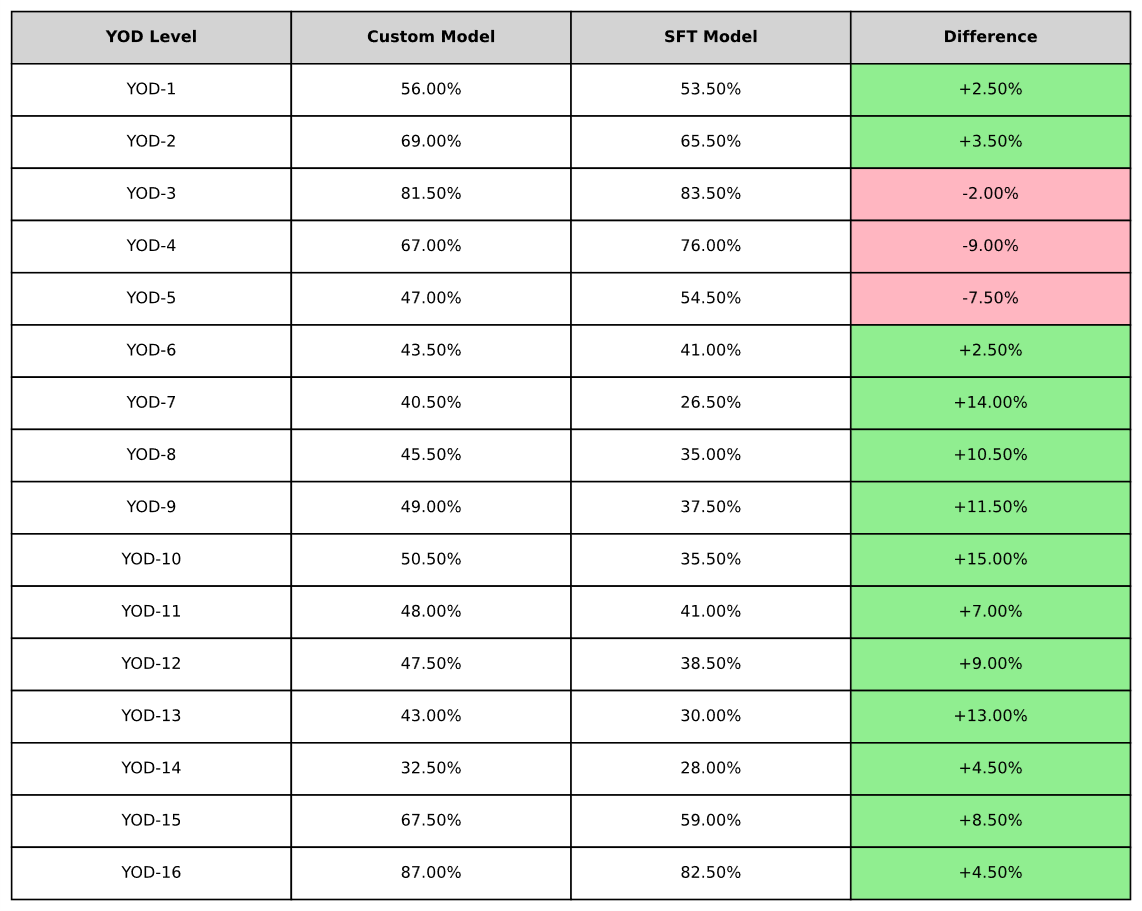}
    \renewcommand{\figurename}{Table}
    \caption{Readability control scores – custom model vs. SFT model.}
    \label{tab:tableyodCom}
    \renewcommand{\figurename}{Figure}
\end{figure}

For readability control, success rates across YOD levels were analyzed as depicted in Table \ref{tab:tableyodCom}  by measuring each model's ability to generate text within ±1.5 levels of the target readability score. 

At lower YOD-levels both models demonstrate good performance, with occasional fluctuations. The custom model demonstrates better performance in YOD-1 (56.00\% vs. 53.50\%) and YOD-2 (69.00\% vs. 65.50\%), while the SFT model exhibits a better performance in YOD-3 (83.50\% vs. 81.50\%), YOD-4 (76.00\% vs. 67.00\%) and YOD-5 (54.50\% vs. 47.00\%).

A notable divergence in performance begins at YOD-7, where the custom model demonstrates stability while the SFT model exhibits a notable decline, as evidenced by a 14\% gap in YOD-7 (40.50\% vs. 26.50\%). In the middle range (YOD 9-12), the custom model consistently demonstrates superior performance, with a margin of 7-15 percentage points, reaching a peak at YOD-10 (50.50\% vs. 35.50\%).

In the higher levels (13-15), the custom model maintains its advantage, with the largest gap occurring at YOD-13 (43.00\% vs. 30.00\%). Both models exhibit a decline at YOD-14 (32.50\% vs. 28.00\%) before a subsequent recovery at YOD-15 (67.50\% vs. 59.00\%). At YOD-16, both models exhibit robust performance, with the custom model demonstrating a slight advantage (87.00\% vs. 82.50\%).

\begin{figure}[H]
    \centering
    \includegraphics[width=1\linewidth]{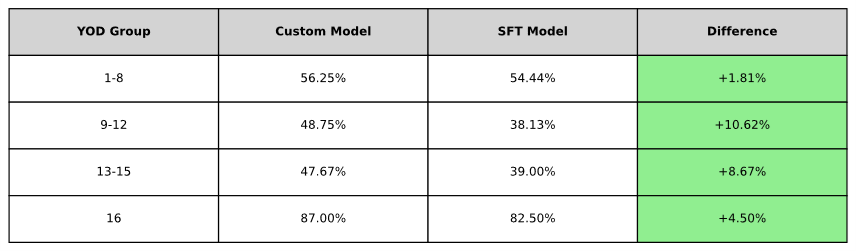}
    \renewcommand{\figurename}{Table}
    \caption{Readability control scores by YOD groups – custom model vs. SFT model.}
    \label{tab:yodGroupsCOm}
    \renewcommand{\figurename}{Figure}
\end{figure}

For readability success rates across YOD groups were analyzed as depicted in Table \ref{tab:yodGroupsCOm} by measuring each model's ability to generate text within ±1.5 levels of the target readability score. 

In the elementary level group (YOD 1-8), both models demonstrate comparable performance, with the custom model showing a slight advantage (56.25\% vs. 54.44\%).

A notable divergence emerges in the high school level group (YOD 9-12), where the custom model maintains stability while the SFT model shows a significant decline, resulting in a substantial performance gap (48.75\% vs. 38.13\%). 
This trend continues into the undergraduate level group (YOD 13-15), where the custom model maintains its superior performance with a marked difference (47.67\% vs. 39.00\%).

At the highest academic/professional level (YOD-16), both models demonstrate robust performance, with the custom model maintaining its advantage but with a narrower margin (87.00\% vs. 82.50\%). This indicates both models' capability to handle complex academic content, though the custom model consistently outperforms across all readability groups.

\section{Limitations}
This section details the various challenges and limitations encountered throughout our study. We identified several key constraints spanning data availability, computational resources, evaluation methodologies, model design, and scope. These limitations underscore the inherent challenges in developing readability-controlled summarization models for Turkish and provide essential context for interpreting our research findings.

\subsection{Data Limitations}
The lack of large-scale Turkish datasets with readability annotations posed a significant challenge for training and evaluation, highlighting the need for specialized data collection efforts. Although readability datasets are available for English, to the best of our knowledge, there are no similar resources for Turkish. This clearly demonstrates the necessity of specially designed data collection and labeling studies to improve the performance of Turkish summarization models.

% Hocam yukarıdaki kısımda ...Turkish are non-existent bu şekilde yazdım.
% ben açık kaynaklı Türkçe için böyle bir dataset görmedim. o yüzden 
% bu kısıma .... similar resources for Turkish are very limited.
% şeklinde değiştirmeli miyim? Çünkü açık kaynaklı olmayabilir ama kapalı 
% kaynaklı olabilir (Tezimde de bu şekilde yazıyor).
%% Yukarıdaki şekilde düzelttim

\subsection{Computational and Resource Limitations}
Developing and fine-tuning large language models for text processing requires substantial computational resources and high-performance GPUs. Access to such hardware is often limited, especially in academic research settings, and the associated costs can be prohibitive.

\subsection{Evaluation Limitations}
Assessing readability in Turkish involves a degree of subjectivity, as interpretations of readability can vary among readers. This subjectivity complicates the evaluation process, particularly when balancing simplification with content accuracy. Quantitative metrics, such as ROUGE, METEOR, and BLEU, may not fully capture this trade-off, highlighting the need for more nuanced evaluation methods that account for linguistic and cultural factors specific to Turkish.

\subsection{Scope Limitations}
This study focuses on specific domains or text types, which may limit the generalizability of its findings across all Turkish content. While the research aims to cover a range of readability levels, it may not fully address the diverse needs of all potential audiences in the Turkish context. Consequently, the applicability of the proposed methodologies may be constrained by the scope of the study.

% belki metnin ilk cümlesindeki specific domains or text type kısımdan
% kastım net anlaşılmayabilir. Benim kullandığım datasetlerin 1'i wikipedia
% summarization dataseti, diğerleri ise haber summarization datasetleri
% sadece formal content olduğu için başka domainlerde iyi performans
% gösteremiyebilir kastım burda bu şekilde.
%% specific domains and text types dedikten sonra bu domain'lerin ne olduğu söyleyebilirsin parantez içinde.

\section{Conclusion}
We constructed a custom dataset and developed a readability-aware summarization model, which we then compared against a supervised fine-tuned (SFT) baseline. Our model, designed with a multi-objective approach, initially exhibited higher losses but demonstrated significant improvements over time. The evaluation highlighted its ability to balance semantic accuracy and readability control, making it a promising alternative for generating summaries with adjustable readability levels.

Furthermore, experimental results show the performance characteristics of the custom-architectured model compared to the supervised fine-tuned (SFT) baseline. Due to computational resource constraints and the current limitations in Turkish language model development, this study focused on optimizing and evaluating a single architectural approach rather than exploring multiple model variants.

Analysis of semantic preservation metrics demonstrates comparable performance between the two models, with minimal variations in effectiveness. The custom model achieved ROUGE-1 scores ranging from 0.345 to 0.666, closely aligned with the SFT model's range of 0.367 to 0.676. This pattern of comparable performance extends to ROUGE-2 and ROUGE-L metrics, with differences remaining within 1.8 percentage points. These results suggest that both architectural approaches maintain similar capabilities in preserving content during the generation process.

The readability control assessment revealed more pronounced differentiation between the approaches. The custom model exhibited better stability in maintaining target readability levels, particularly evident in the challenging mid-range YOD levels (9-12), where it consistently outperformed the SFT model by 7-15 percentage points. This stability is particularly noteworthy given the inherent complexity of simultaneously controlling readability while maintaining semantic coherence.
Examination of the learning dynamics revealed distinct patterns between the two approaches. While the SFT model demonstrated a gradual, consistent improvement in training loss from 0.24 to 0.156, the custom model exhibited a more aggressive learning curve, rapidly decreasing from 4.60 to 0.28. This suggests that the custom architecture's more complex training process, involving three simultaneous neural networks, enables more effective feature learning despite higher initial losses and limited computational resources.

The evaluation framework, incorporating a ±1.5 YOD tolerance range, provides a robust methodology for assessing controlled text generation systems while acknowledging the probabilistic nature of neural language models. The custom model's ability to maintain stability in challenging YOD levels while achieving comparable semantic preservation metrics represents a significant advancement in controlled text generation capabilities, particularly within the constraints of available Turkish language models and computational resources.

These findings contribute to the broader understanding of controlled text generation and establish a foundation for future research directions. The demonstrated success of the implemented architectural approach, despite hardware limitations and language-specific constraints, suggests potential benefits in exploring sophisticated architectural designs for controlled generation tasks. Future research directions might explore the integration of additional control mechanisms or the application of this architecture to other text generation tasks requiring fine-grained control over output characteristics, dependent on the availability of more advanced computational resources and improved large language models.

\newpage
\label{sec:headings}
\bibliographystyle{unsrtnat}

\end{document}